# Hostile Intent Identification by Movement Pattern Analysis: Using Artificial Neural Networks

Souham Biswas
J.K. Institute of Applied Physics & Technology
University of Allahabad
Allahabad, India
souhambiswas@outlook.com

Manisha J. Nene
Dept. of Applied Mathematics and Computer Engineering
Defence Institute of Advanced Technology, Defence
R&D Organization, Ministry of Defence
Pune, India
mjnene@diat.ac.in

*Abstract*—In the recent years, the problem of identifying suspicious behavior has gained importance and identifying this behavior using computational systems and autonomous algorithms is highly desirable in a tactical scenario. So far, the solutions have been primarily manual which elicit human observation of entities to discern the hostility of the situation. To cater to this problem statement, a number of fully automated and partially automated solutions exist. But, these solutions lack the capability of learning from experiences and work in conjunction with human supervision which is extremely prone to error. In this paper, a generalized methodology to predict the hostility of a given object based on its movement patterns is proposed which has the ability to learn and is based upon the mechanism of humans of "learning from experiences". The methodology so proposed has been implemented in a computer simulation. The results show that the posited methodology has the potential to be applied in real world tactical scenarios.

## I. Introduction

One of the most daunting tasks pertaining to the defence forces of a country has always been the effective identification and elimination of threats which interfere with the interests or the security of the nation. In situations of conflict, the fallibility of human judgement caused by stress or any such factor in determining the hostility of a given target can prove to be fatal and might result in considerable loss of resources. Therefore, a need to automate the same is highly desirable.

The term "hostility" is inherently multifarious. The meaning depends upon the observer. This implies that, there are a considerable number of variables pertaining to this characteristic, if viewed analytically. Every day, new attack techniques are observed and defence tactics are being innovated. It is impossible to define an all-encompassing set of parameters or variables which would successfully quantify "hostility" in a general sense. However, there is one parameter of hostility pertaining to the object in question that spans over the others and is potentially impervious to the nature of the observer; the location/existence of the object under observation. The location of a given object can be grilled to obtain a multitude of characteristics from which certain behavioural traits can be extracted and analysed. Although, a perfect analytical solution to this problem statement is far-fetched as of now, a more promising approach is to incorporate the way humans try to solve this problem into a machine; to include the element of "intuition". The approach posited here draws on the fact that the human mind is a continuously evolving palimpsest of neurons, which adapt and evolve to any situation. Hence, the deployment of artificial neural networks; even more for their reputation for being extremely fault tolerant which is of paramount importance when considering such problems with high margins for error.

*Background and related work:*

During the literature survey, it was observed that a small set of analytical approaches do exist [1], [5], which address this problem. The most prominent of which, is the US Patent termed "Detection of Hostile Intent from Movement Patterns" [1]. But, what these approaches lack is the trait to adapt according to situation. It is not possible to discretely classify a given behaviour as "hostile" or "not hostile". Therefore, it is imperative that the system intended to make that classification, *learn* how to do so by itself and adapt in pace with the constantly evolving attack/defence tactics.

Presented in this paper is an approach which seeks to make for the shortcomings faced by the present systems. The paper is organized as follows: Section II describes the various parameters and assumptions involved and basis of the methodology. Section III enumerates the actual methodology, critical parameters, algorithms and the processes involved. In Section IV, simulation results have been elucidated. Finally, Section V summarises and concludes the proposed work and mentions the scope for future work.

## II. Proposed Work

### A. Basis of Hostility Detection

One of the most prominent characteristics of the human brain is the tendency to correlate between new information and previous "experiences" to draw conclusions [7]. The degree of this correlation and the subsequent processing of the same allow us to make fuzzy predictions [6] or in one way, form an *intuition*. The notion of hostility in general, lies in previous experiences of such situations endured by an individual. When presented with a scenario for hostility detection, the brain tries to discern the degree of similarity between the new situation and a catalogue of "hostile" labelled situations previously encountered. A high similarity calls for evasive measures. To summarize, one takes steps to ensure that the sequence of events which led to the previously sustained events of hostility do not repeat. To model this as an automated solution, we consider neural networks. We follow a similar process of training the network that is; an expansive dataset of known hostile situations is made incident on the network. As the training proceeds, the network tends to form its own notions for enumerating hostilities. In other words, an artificial sense of *intuition* is formed.

*B. Assumptions & Parameters Involved*

To parameterize a given situation for neural network training [4], we consider the locations of the objects. This quantity maybe in parametric, polar or any other co-ordinate form. The definitions of a few prominent terminologies are given below-

- **Area of Observation:** It is the physical region which is being monitored for hostile activities. Practically, this can translate to a given region on the shore, the range of a radar, etc.

- **Object:** Any entity inside the area of observation which will be subject to probation for determination of *hostility* is termed as "object".

- **Hostility:** It is the probability that an object will commit an act of hostility in the immediate future.

Other parameters derived from the locations of objects like the speed and direction of object, density of objects in a given area etc. may also be computed and added as inputs to the neural network. The shape of the area of observation does not pose any constraint in location determination.

### III. METHODOLOGY

The system will take inputs as the locations of the multiple objects inside the area of observation in the form of X and Y coordinates. The neural network being utilized will be a 2-layer feed forward network [8] with sigmoid function (1) as the activation function.

$$f_{sig}(x) = \frac{1}{1+e^{-x}}. \quad (1)$$

The datasets involved in training are of the following types-

- *Raw Dataset* – This contains the records of locations of all the objects in the area of observation and their corresponding probabilities of hostility.

- *Normalized Dataset* – This is the dataset which is actually used to train the neural network. Normalized Dataset is obtained by generating all the permutations of the raw dataset.

*System Variables and Relations-*

- $N$ : Number of objects inside the area of observation.

- $M^k$ : Number of entries in raw dataset having dataset index "$k$" ($k^{th}$ raw dataset).

- $K$ : Number of training datasets.

- $M'^k$ : Number of entries in normalized dataset having dataset index "$k$" ($k^{th}$ raw dataset).

- $X_v^{u^k}$ : X coordinate of object having index "$v$" in $k^{th}$ raw dataset at observation index "$u$".

- $X'^{u^k}_v$ : X coordinate of object having index "$v$" in $k^{th}$ normalized dataset at observation index "$u$".

- $Y_v^{u^k}$ : Y coordinate of object having index "$v$" in $k^{th}$ raw dataset at observation index "$u$".

- $Y'^{u^k}_v$ : Y coordinate of object having index "$v$" in $k^{th}$ normalized dataset at observation index "$u$".

- $\Omega_v^{u^k}$ : Probability of hostility of object having index "$v$" in $k^{th}$ raw dataset at observation index "$u$".

- $\Omega'^{u^k}_v$ : Probability of hostility of object having index "$v$" in $k^{th}$ normalized dataset at observation index "$u$".

- $P_v^{u^k}$ : This denotes the location of object having index "$v$" at observation index "$u$" at the $k^{th}$ raw dataset index.

- $P'^{u^k}_v$ : This denotes the location of object having index "$v$" at observation index "$u$" at the $k^{th}$ normalized dataset index.

- $A_k^u$ : Locations of all objects (sets of X-Y coordinates) in $k^{th}$ raw dataset at observation index "$u$".

- $A'^u_k$ : Locations of all objects (sets of X-Y coordinates) in $k^{th}$ normalized dataset at observation index "$u$".

- $B_k^u$ : Probabilities of hostility of all objects (sets of $\Omega_v^{u^k}$ values) in $k^{th}$ raw dataset at observation index "$u$".

- $B'^u_k$ : Probabilities of hostility of all objects (sets of $\Omega'^{u^k}_v$ values) in normalized dataset at $k^{th}$ observation index "$u$".

- $T_k^u$ : Raw training data having observation index "$u$" at $k^{th}$ training dataset.

- $T'^u_k$ : Normalized training data having observation index "$u$" at $k^{th}$ training dataset.

- $D_r$ : Raw training dataset.

- $D_n$ : Normalized training dataset.

- $Q^k$ : $k^{th}$ raw dataset.

- $Q'^k$ : $k^{th}$ normalized dataset.

- $L^k$ : Dataset containing location data of all objects in the area of observation with raw dataset index "$k$" ($k^{th}$ raw dataset).

- $H^k$ : Dataset containing hostility probability data of all objects in the area of observation with raw dataset index "$k$" ($k^{th}$ raw dataset).

- $L'^k$ : Dataset containing location data of all objects in the area of observation with normalized dataset index "$k$" ($k^{th}$ normalized dataset).
- $H'^k$ : Dataset containing hostility probability data of all objects in the area of observation with normalized dataset index "$k$" ($k^{th}$ normalized dataset).

*Object index* is a number assigned to each of the objects inside the area of observation for uniquely identifying them.

*Relations* –

The mathematical relations between the variables mentioned previously are enumerated as follows.

$$P_v^{u^k} = \{X_v^{u^k}, Y_v^{u^k}\} \quad (2)$$

$$A_k^u = \{P_v^{u^k} : v \in [1,N]\} \ \forall \ u \in [1,M^k], k \in [1,K] \quad (3)$$

$$L^k = \{A_k^u : u \in [1,M^k]\} \ \forall \ k \in [1,K] \quad (4)$$

$$B_k^u = \{\Omega_v^{u^k} : v \in [1,N]\} \ \forall \ u \in [1,M^k], k \in [1,K] \quad (5)$$

$$H^k = \{B_k^u : u \in [1,M^k]\} \ \forall \ k \in [1,K] \quad (6)$$

$$T_k^u = \{A_k^u, B_k^u\} \ \forall \ k \in [1,K], u \in [1,M^k] \quad (7)$$

$$Q^k = \{L^k, H^k\} \ \forall \ k \in [1,K] \quad (8)$$

$$D_r = \{Q^k : k \in [1,K]\} \quad (9)$$

$$P'^{u^k}_v = \{X'^{u^k}_v, Y'^{u^k}_v\} \quad (10)$$

$$A'^{u}_k = \{P'^{u^k}_v : v \in [1,N]\} \ \forall \ u \in [1,M'^k], k \in [1,K] \quad (11)$$

$$L'^k = \{A'^{u}_k : u \in [1,M'^k]\} \ \forall \ k \in [1,K] \quad (12)$$

$$B'^{u}_k = \{\Omega'^{u^k}_v : v \in [1,N]\} \ \forall \ u \in [1,M'^k], k \in [1,K] \quad (13)$$

$$H'^k = \{B'^{u}_k : u \in [1,M'^k]\} \ \forall \ k \in [1,K] \quad (14)$$

$$T'^{u}_k = \{A'^{u}_k, B'^{u}_k\} \ \forall \ k \in [1,K], u \in [1,M'^k] \quad (15)$$

$$Q'^k = \{L'^k, H'^k\} \ \forall \ k \in [1,K] \quad (16)$$

$$D_n = \{Q'^k : k \in [1,K]\} \quad (17)$$

$$M'^k = M^k \times (N!) \quad (18)$$

### A. Procurement of Training Data

Initially, the set $D_r$ is to be generated which is basically the *raw dataset* as previously explained.

TABLE I. TABLE REPRESENTATION OF $L^1$ SET

| Sr. No. | Object Locations | | | | | |
|---|---|---|---|---|---|---|
| | A1 | B1 | A2 | B2 | A3 | B3 |
| 1. | 234 | 874 | 214 | 856 | 764 | 214 |
| 2. | 045 | 698 | 102 | 523 | 154 | 601 |
| 3. | 487 | 035 | 924 | 157 | 245 | 682 |
| 4. | 147 | 256 | 651 | 654 | 213 | 746 |

a. Here $k = 1$ for $L^k$

Table I is a tabular illustration of sample $L^1$, since this is the first dataset, $k = 1$. Here, the cell at index (2, $A3$) can be represented as $X_3^{2^1}$; the same can be extended to the other cells. This example dataset assumes there are only 3 objects in the area of observation. Similarly, we can have multiple training datasets.

TABLE II. TABLE REPRESENTATION OF $L^2$ SET

| Sr. No. | Object Locations | | | | | |
|---|---|---|---|---|---|---|
| | A1 | B1 | A2 | B2 | A3 | B3 |
| 1. | 568 | 248 | 278 | 698 | 421 | 297 |
| 2. | 354 | 014 | 685 | 032 | 682 | 413 |
| 3. | 570 | 694 | 724 | 031 | 824 | 246 |

b. Here $k = 2$ for $L^k$

Table II illustrates another dataset involved in training. Note that the two datasets are mutually independent and merely represent the log of locations of the multiple objects in the area of observation when an event of hostility had been previously sustained. Tables III and IV illustrate the tabular representations of the observed hostility probabilities ($H^k$) of the objects in the area of interest for each of the datasets with $k = 1$ and $k = 2$ respectively.

TABLE III. TABLE REPRESENTATION OF $H^1$ SET

| Sr. No. | Object Hostility Probabilities | | |
|---|---|---|---|
| | A1 | A2 | A3 |
| 1. | 0.00 | 0.00 | 1.00 |
| 2. | 0.00 | 1.00 | 0.00 |
| 3. | 0.00 | 0.00 | 1.00 |
| 4. | 1.00 | 0.00 | 1.00 |

c. Here $k = 1$ for $H^k$

TABLE IV. TABLE REPRESENTATION OF $H^2$ SET

| Sr. No. | Object Hostility Probabilities | | |
|---|---|---|---|
| | A1 | A2 | A3 |
| 1. | 1.00 | 0.00 | 0.00 |
| 2. | 1.00 | 0.00 | 0.00 |
| 3. | 0.00 | 0.00 | 1.00 |

d. Here $k = 2$ for $H^k$

The probabilities in Table III are only "0" or "1" because the network will undergo supervised training. The cell at index (3, $A1$) in Table IV can be represented as $\Omega_1^{3^2}$; the same can be extended to the other cells. The system variables defined previously are illustrated in the context of the present example in the succeeding text.

- $N = 3$
- $K = 2$
- $M^1 = 4$
- $M^2 = 5$

- $L^1$ = Table I
- $L^2$ = Table II
- $H^1$ = Dataset containing hostility probability data of all objects in the area of observation with raw dataset index "1" (Table III). Similarly, $H^2$ is defined.
- $A_2^3$ = 3rd row of Table II.
- $B_1^2$ = 2nd row of Table III.
- $D_r$ = Collection of all tables I-IV organized as {(Table I, Table III), (Table II, Table IV)}

Similarly, the other system variables can be computed. In practical application, this data can be obtained by analysing previous events of hostility sustained. The $D_r$ set so generated cannot be used to train the neural network yet. It has to be subjected to *normalization* to get $D_n$ (normalized dataset) which will be used to train the neural network.

### B. Generation of Normalized Training Data

Normalization refers to generation of all permutations of the sets $L^k$ & $H^k$ for all $k$ from 1 to $K$. This process is important because, a hostile object need not be assigned the same object index every time it is inside the area of observation. For example, suppose the neural network is trained using $D_r$ and that in the dataset, for some $n, u$ and $k$, $\Omega_n^{u^k} = 1.00$. Correspondingly, the network is trained to output $\Omega_n^{u^k} = 1.00$ whenever the input is $A_k^u$. Here, it is evident that the object with index $n$ is hostile. But suppose in the future, the same object is assigned an object index of $n'$; then, the system will *fail* to identify successfully this hostile object as it has been trained to identify the hostile traits of object at index $n$ and not at $n'$. Although, if the system is also trained with all the permutations of $A_k^u$ and $B_k^u$ as input and output respectively, the system will always identify the hostile object irrespective of the object index assigned to it. To explain the normalization process, consider a scenario with $N = 2, K = 1, M^1 = 2$.

$$\therefore A_1^1 = \{P_1^{1^1}, P_2^{1^1}\}$$
$$A_1^2 = \{P_1^{2^1}, P_2^{2^1}\}$$

$$B_1^1 = \{\Omega_1^{1^1}, \Omega_2^{1^1}\}$$
$$B_1^2 = \{\Omega_1^{2^1}, \Omega_2^{2^1}\}$$

$\therefore$ Generating all permutations of $A_1^1, A_1^2, B_1^1$ and $B_1^2$

$$A'^1_1 = \{P_1^{1^1}, P_2^{1^1}\}$$
$$A'^2_1 = \{P_2^{1^1}, P_1^{1^1}\}$$

$$A'^3_1 = \{P_1^{2^1}, P_2^{2^1}\}$$
$$A'^4_1 = \{P_2^{2^1}, P_1^{2^1}\}$$

Similarly, generate normalized dataset for set $H^1$

$$M'^1 = M^1 \times (N!) \quad \text{[From (18)]}$$
$$\Rightarrow M'^1 = 2 \times (2!) = 8$$

For some $k \in [1, K], u \in [1, M'^k]$

$$T'^u_k = \{A'^u_k, B'^u_k\} \quad \text{[From (15)] (19)}$$

Eq. 19 is the normalized training data at observation index "$u$" as previously stated in Section II, B. In this, $A'^u_k$ is input data to the neural network and $B'^u_k$ is the set of target outputs. The derivation of $D_n$ is as follows –

$$L'^k = \{A'^u_k : u \in [1, 8]\} \forall k \in [1, 1] \quad \text{[From (12)]}$$
$$H'^k = \{B'^u_k : u \in [1, 8]\} \forall k \in [1, 1] \quad \text{[From (14)]}$$
$$Q'^k = \{L'^k, H'^k\} \forall k \in [1, 1] \quad \text{[From (16)]}$$
$$D_n = \{Q'^k : k \in [1, 1]\} \quad \text{[From (17)]}$$

### C. Neural Network Training

The structure of the neural network to be trained by $D_n$ is illustrated in Fig. 1.

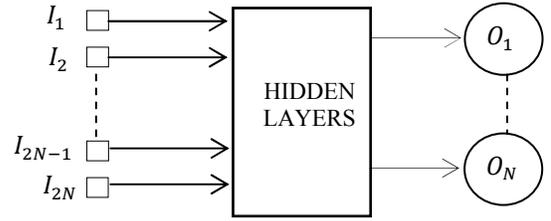

Fig. 1 Structure of Neural Network: A 2-Layer Feed Forward Network with 2N input neurons and N output neurons

As shown in Fig. 1 the set $I$ is defined as-

$$I = \{I_v : v \in [1, 2N]\}$$

Hence, $I$ represents the set of input nodes and for an object with index $v$, the set $p_v$ is defined as –

$$p_v = \{I_{2v-1}, I_{2v}\}$$

The set $p_v$ represents the location of object having index $v$ as a set of X and Y co-ordinates. Similarly, $O_v$ is the *hostility* of object having index $v$. While training, for a given $Q'^k$, we set $p_v = P'^{u^k}_v$ and $O_v = \Omega'^{u^k}_v$ and cycle the value of $u$ from 1 to $M'^k$. In each iteration, the system is trained using backpropagation and gradually the certitude with which the system predicts the hostility probability of each object increases. We repeat this process for each training dataset i.e. for all $k \in [1, K]$. But in each dataset, only 70% should be used for training, 20% for validation and the remaining 10% for testing purposes. The distribution of the data amongst these three groups has to be random.

### D. Validation of Neural Network

The process of validation is carried out so as to determine when to stop training and to avoid over-fitting. At each iteration, *error* is calculated from the validation data, the formula of which is given in Eq. 38.

$$E = \sum_{j=1}^{m} \int [y_j(x, w) - t_j]^2 p(x, t_j). dx \quad (20)$$

Eq. 20 is used, as the network is a feed-forward network which is trained using back-propagation [2]. The network here, is a set of functional mappings $y_j(x, w)$ [3], which relate an input $x$ with a given set of bias weights $w$ whose values are obtained through minimizing $E$. Here, the joint probability density functions for the training data are given by $p(x, t_j)$ where $j = 1,2,\ldots,N$ corresponds to each of the output neurons, $y_j$ is the output of neuron $j$ and $t_j$ is the target output for that neuron. Initially the error decreases and the gradient of the error decrease rate changes until approaching zero. Training stops when generalization stops improving network performance as measured from the validation data.

*E. Deployment of Neural Network*

The flowchart to illustrate the system working is given in Fig. 2.

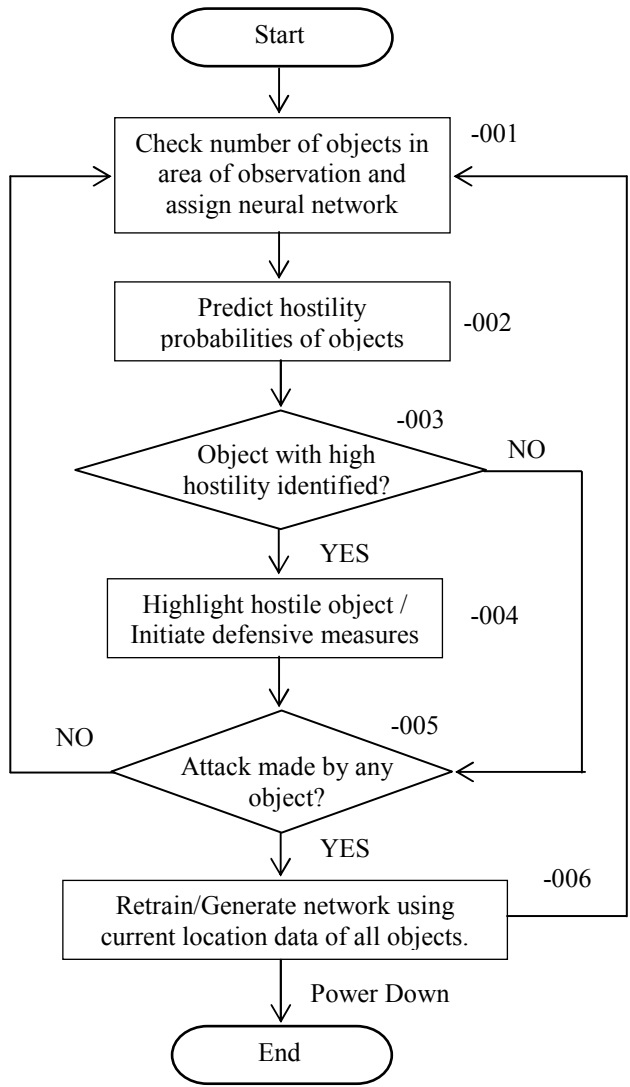

Fig. 2 Process Flowchart for Hostility Prediction System

A given neural network can cater to only a fixed number of objects in the area of observation. Therefore, multiple neural networks having different number of inputs must be trained and generated before pragmatic deployment. For example, first, a neural network (having 8 inputs) maybe generated to analyze situations when only 4 objects are present in the area of observation; now, if the number of objects changes to 5, another neural network having 10 inputs will be needed. In node 001, the corresponding network is chosen. Subsequently in node 002, hostility probabilities of all the objects are calculated; now if an object with alarming hostility is identified (003), defensive measures should be taken to avoid any casualty. Finally, if the system *fails* to warn of an impending hostile event, then the network will retrain itself and *learn* from the experience after the hostile situation has subsided (006); much like the way humans learn from experiences. Therefore, a similar attack could be prevented in the future. This is indeed a drawback and is caused due to incomplete training. This is the reason why the network is trained with an expansive dataset of known hostile situations before deployment.

IV. SIMULATION & RESULTS

The proposed system has been simulated and implemented using MATLAB and C# on MonoGame® Framework. The neural network was generated in MATLAB. The simulation involves a front-end C# GUI application simulating a maritime radar environment running in parallel with MATLAB in the background in which the actual neural network computation and live-time training is taking place.

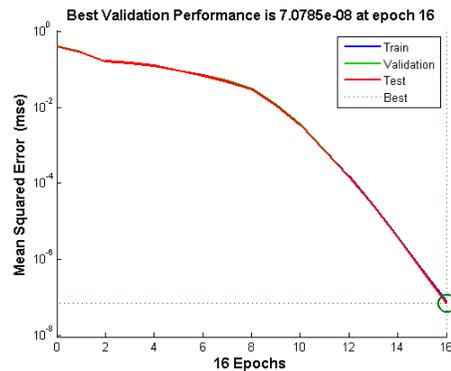

Fig. 3 Confusion Matrix Pertaining to Testing Phase of Neural Network

Fig. 4 Training Performance Measure of Neural Network

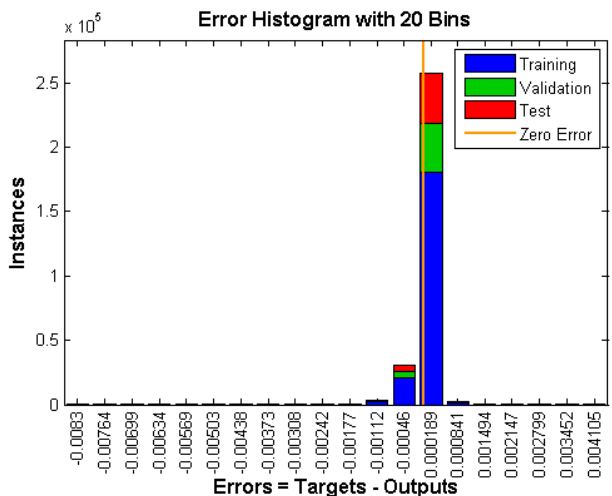

Fig. 5 Error Histogram of Training

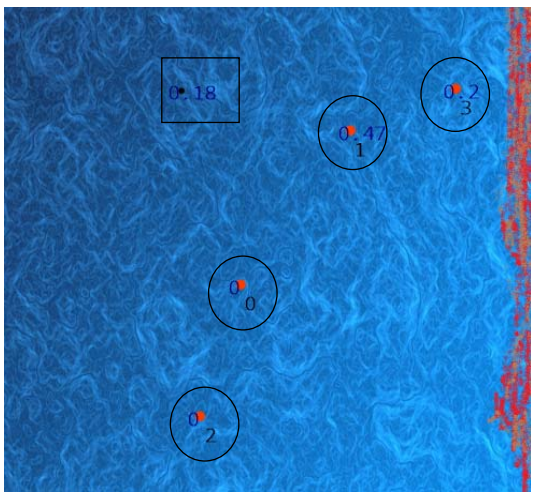

Fig. 6 System Simulation

For simulation purposes, a scenario with system variables defined as $N = 5$, $K = 1$, $M^1 = 494$ was considered. Fig. 3, 4 & 5 show characteristics of the neural network right after training and testing. Shown in Fig. 3 is the confusion matrix which maps the outputs of the neural network to that of the actual output over test data (10% of the dataset). By observing the diagonal, it is evident that the neural network has correctly predicted the hostilities of all objects for all inputs. Fig. 4 illustrates how the mean squared error of the network decreases as training proceeds. We can derive from Fig. 4 that the network starts to produce proper results after 16 training iterations. Fig. 5 basically shows the error range in which most of the network outputs lie, over the input data. Here, as most of the outputs lie in the error range of around 0.000169 which is negligible, we can say that the network outputs hostility probabilities with sufficient certitude. Fig. 6 is a screenshot of the actual simulation. Here, each dot represents an object inside the area of observation. The dot enclosed in the box is controllable by the user. To the extreme right, there is a landmass which is to be protected. The encircled dots have two numbers associated with them placed one on top of the other; the one below, represents object index and the other above, is the *probability of hostility* of the object. It was observed that the patterns of attack used to train the neural network, were successfully highlighted in this implementation whenever an attack with a similar pattern was made. Moreover, if the user performed an attack, then in the future, a similar kind of attack made by the user was automatically highlighted, which exemplifies the *learning* ability of the proposed system. The simulation process was carried out over a number of scenarios with different system variables and similar results were obtained.

## V. CONCLUSION

A solution has been proposed which can serve as the basis for fully automating the process of hostile intent detection. The system takes inputs as only the locations of the objects; therefore, it can be directly deployed in conjunction with radar systems or other visual surveillance systems. Apart from the location of the target, the incorporation of other factors specialized to the domain of application of this system has the potential to yield results with greater certitude. However, the system presented in this paper incorporates only the locations of the object so as to maintain a subtle degree of generality. Therefore, a framework has been presented which can be specialized to encompass different domains of hostile object detection.

The framework posited in this paper has many applications in fields apart from maritime surveillance and radar systems. The system can be deployed in cyber-security domains for analysis and detection of malicious data packets. The very fact that the system has the ability to *learn* and adapt to new tactics by accepting data pertaining to the object *not* limited to only the location of which, greatly expands the domain of application of this methodology.